%% file: main.tex
\documentclass[sigconf]{acmart}
\include{preamble}

\newlength{\whilewidth}
\algdef{SE}[parWHILE]{parWhile}{EndparWhile}[1]
  {\parbox[t]{\dimexpr\linewidth-\algmargin}{%
     \hangindent\whilewidth\strut\algorithmicwhile\ #1\ \algorithmicdo\strut}}{\algorithmicend\ \algorithmicwhile}%

\AtBeginDocument{%
  \providecommand\BibTeX{{%
    \normalfont B\kern-0.5em{\scshape i\kern-0.25em b}\kern-0.8em\TeX}}}

\usepackage{xspace}
\usepackage{threeparttable}

\begin{document}

\title[\resizebox{4.5in}{!}{Dynamic Co-Optimization Compiler: Leveraging Multi-Agent Reinforcement Learning for Enhanced DNN Accelerator Performance}]{ Dynamic Co-Optimization Compiler: Leveraging Multi-Agent Reinforcement Learning for Enhanced DNN Accelerator Performance}


\author{Arya Fayyazi}
\affiliation{%
  \institution{University of Southern California}
  \city{Los Angeles}
  \state{California}
  \country{USA}
}
\email{afayyazi@usc.edu}

\author{Mehdi Kamal}
\affiliation{%
  \institution{University of Southern California}
  \city{Los Angeles}
  \state{California}
  \country{USA}
}
\email{mehdi.kamal@usc.edu}


\author{Massoud Pedram}
\affiliation{%
  \institution{University of Southern California}
  \city{Los Angeles}
  \state{California}
  \country{USA}
}
\email{pedram@usc.edu}

\renewcommand{\shortauthors}{Fayyazi, et al.}

\newcommand{\alg}{{DCOC}\xspace}

\begin{abstract}
This paper introduces a novel Dynamic Co-Optimization Compiler (DCOC), which employs an adaptive Multi-Agent Reinforcement Learning (MARL) framework to enhance the efficiency of mapping machine learning (ML) models, particularly Deep Neural Networks (DNNs), onto diverse hardware platforms. DCOC incorporates three specialized actor-critic agents within MARL, each dedicated to different optimization facets: one for hardware and two for software. This cooperative strategy results in an integrated hardware/software co-optimization approach, improving the precision and speed of DNN deployments. By focusing on high-confidence configurations, DCOC effectively reduces the search space, achieving remarkable performance over existing methods. Our results demonstrate that DCOC enhances throughput by up to 37.95\% while reducing optimization time by up to 42.2\% across various DNN models, outperforming current state-of-the-art frameworks.
\end{abstract}
\vspace{-2mm}

\vspace{-2mm}

\maketitle
\vspace{-2mm}

\section{Introduction}

The computational demands and complexity of deep neural networks (DNNs) have surged with their expanding applications across various industries such as autonomous driving, medical imaging, and natural language processing \cite{hossain2023computational,Ashrafi2024.06.26.24309553, diagnostics13020179,202410.1684,razmara2024feverdetectioninfraredthermography}. Traditionally, performance optimization in frameworks like TensorFlow \cite{abadi2016tensorflow} and PyTorch \cite{paszke2019pytorch} has relied on hand-optimized kernels like NVIDIA's cuDNN and Intel's MKL. However, these kernels struggle to adapt to the evolving demands of modern neural architectures \cite{ahn2020chameleon}.

To address this, automated compilation frameworks like TVM \cite{chen2018tvm}, TensorComprehensions \cite{vasilache2018tensor}, and AutoTVM \cite{chen2018tvm} have emerged, shifting from static, hand-tuned optimization to dynamic, algorithm-driven approaches. For instance, AutoTVM uses boosted trees \cite{chen2016xgboost} to efficiently explore the configuration space of neural network code. Despite these advances, optimizing complex models still requires substantial resources and compilation time, highlighting ongoing challenges in current optimization methodologies. Reinforcement learning (RL) has shown promise in exploring design search spaces and optimizing tasks \cite{zhang2022harl,wang2022automating,bakshi2023computationally,ahn2020chameleon}, with multi-agent RL offering advantages like transfer learning and insights into interconnected decision-making systems \cite{du2021survey}.

In this work, we propose an adaptive multi-agent reinforcement learning-based hardware/software co-optimization compiler called \alg. Our compiler employs a multi-agent reinforcement learning (MARL) algorithm to concurrently optimize the mapping of the DNN model and the accelerator architecture. The use of MARL effectively manages the expansive search space of hardware and software configurations, leveraging the collaborative strengths of multiple agents \cite{yu2022surprising}, which is crucial due to the significant impact of hardware-software interplay on overall performance.

\alg features three specialized agents: two focusing on software and DNN configurations, and one optimizing the accelerator architecture. This actor-critic MARL approach enables an integrated and holistic optimization process with granular control over optimization strategies. The framework efficiently narrows the vast search space by employing a soft threshold to select high-confidence configurations, concentrating on the most promising candidates, thus reducing computational overhead and exploration time.

Utilizing a Centralized Training with Decentralized Execution (CTDE) strategy, \alg allows adaptive adjustments to the changing requirements of DNN workloads. We assess the efficacy of \alg on a set of DNN models with various architectures, comparing it with leading state-of-the-art approaches. The results demonstrate substantial improvements in throughput, showcasing the potential of MARL-based co-optimization in advancing compilation technologies.

\section{Background}
\label{sec:prelim}

\subsection{CTDE in MARL}
\label{subsec:ACMARL}

Multi-agent reinforcement learning (MARL) extends traditional reinforcement learning by introducing multiple agents within a shared environment, necessitating methods to coordinate strategies towards collective objectives. The Centralized Training with Decentralized Execution (CTDE) framework balances collective intelligence during training with agent autonomy during execution \cite{lyu2021contrasting}.

In CTDE, agents have access to global state information during centralized training, allowing them to learn cohesive policies that consider all agents' states and actions. This approach effectively addresses cooperative and competitive tasks. Proximal Policy Optimization (PPO) \cite{schulman2017proximal} has been adapted to this framework, resulting in Multi-Agent PPO (MAPPO) \cite{yu2022surprising}.

The MAPPO learning process involves three key components:

\begin{itemize}
\item \textbf{Critic learning}: Enhances the centralized value function:
\begin{equation}
\phi_{k+1} = \arg\min_{\phi} \frac{1}{|D_{k}|T} \sum_{\tau \in D_{k}} \sum_{t=0}^{T} \left( V_{\phi}(\mathbf{o}_t, \mathbf{s}_t, \mathbf{u}_t) - \hat{R}_t \right)^2
\end{equation}
\item \textbf{Generalized Advantage Estimation (GAE)}: Gauges the quality of the current action relative to the baseline:
\begin{equation}
A_{t} = \sum_{t=0}^{\infty} (\gamma \lambda)^{t} \delta^{V}_{t+1}
\end{equation}
\item \textbf{Policy learning}: Updates the policy using the estimated advantage:
\begin{equation}
\begin{aligned}
L(\mathbf{o}, \mathbf{s}, \mathbf{u}, \mathbf{u}', \theta_{k}, \theta) = & \min \bigg( \frac{\pi_{\theta}(\mathbf{u}|\mathbf{o})}{\pi_{\theta_{k}}(\mathbf{u}|\mathbf{o})} A^{\pi_{\theta_{k}}}(\mathbf{o}, \mathbf{s}, \mathbf{u}'), \\
& \text{clip}\left( \frac{\pi_{\theta}(\mathbf{u}|\mathbf{o})}{\pi_{\theta_{k}}(\mathbf{u}|\mathbf{o})}, 1 - \epsilon, 1 + \epsilon \right) A^{\pi_{\theta_{k}}}(\mathbf{o}, \mathbf{s}, \mathbf{u}') \bigg)
\end{aligned}
\end{equation}
\end{itemize}

Empirical studies show that MAPPO can rival and sometimes outperform competitive off-policy algorithms \cite{yu2022surprising}, leading us to adopt MAPPO as our optimization algorithm in MARL.

\subsection{Workflow for DNN Compilers}
\label{subsec:wflow_DNN}

Efficiently compiling DNN models into machine code is crucial for performance across various hardware platforms. Compiler workflows optimize the DNN model in several stages \cite{allen2002optimizing}. Initially, the compiler's frontend performs general optimizations without considering specific hardware. Subsequently, backend optimizations tailor the code based on the target hardware's characteristics.

Recent advancements like AutoTVM have introduced a fine-tuning step that uses hardware performance feedback to explore various configuration settings ("knobs") and determine the most effective code configuration. The search for the optimal configuration ($\Theta^*$) is expressed as:
\[
\Theta^* = \underset{\Theta \in D}{\text{argmax}} \left\{ f\left[\tau(\Theta)\right] \right\},
\]
where \( f \) represents the performance metric, \( \Theta \) denotes a specific set of knob settings, and \( D \) is the allowed configuration space. The compiler's goal is to discover the settings that maximize performance on the target hardware.

\subsection{Related Work}
\label{subsec:prior_work}

In the domain of DNN compilers, adaptive and efficient auto-tuning mechanisms are essential for optimizing performance on varying hardware architectures \cite{rieber2022hw,10.1145/3665314.3670843,ryu2022one,lu2022semi,10.1145/3649476.3658736,dhakal2022slice,azizi2024efficientnoisemitigationenhancing}. AutoTVM \cite{chen2018tvm} utilizes a machine learning-based cost model, specifically XGBoost \cite{chen2016xgboost}, but involves a large search space and substantial measurement overhead.

MetaTune \cite{ryu2021metatune} addresses these limitations by leveraging meta-learning to quickly adapt to new optimization spaces, demonstrating improved inference times. Glimpse \cite{ahn2022glimpse} integrates mathematical embeddings of hardware specifications, guiding the search algorithm towards higher performance subspaces using Bayesian optimization.

CHAMELEON \cite{ahn2020chameleon} employs reinforcement learning to minimize invalid configurations and costly hardware measurements, reducing optimization time and improving inference times compared to AutoTVM. NaaS \cite{zhou2022towards} jointly optimizes neural network architectures and hardware accelerators, parameterizing both in a unified search space using PyGlove. PRIME \cite{kumar2021data} introduces a data-driven method for designing hardware accelerators, using logged simulation data to construct a robust surrogate model, significantly reducing simulation time.

Comparing these methods, AutoTVM lays the groundwork for machine learning-based auto-tuning. MetaTune and Glimpse advance this with meta-learning and hardware-aware strategies, respectively. CHAMELEON enhances the reinforcement learning approach with improved sampling methods. NaaS and PRIME introduce accelerator co-design and offline data-driven optimization but lack reinforcement learning, making them slower in compilation times.

Our approach differs from existing methods like CHAMELEON, Glimpse, and MetaTune—which reduce compilation times through various optimization techniques—and co-design methods like NaaS and PRIME—which enhance system throughput. Our analysis highlights the advantages of MARL in efficiently navigating the design space in modern DNN applications, capturing hardware and software interdependencies to achieve optimal performance and throughput in real-world applications.


\section{Proposed DCOC}
\label{sec:DCOC}

Existing systems often fail to optimize both hardware and software components of DNNs simultaneously, focusing primarily on software parameters, while co-design approaches are too time-consuming to compile. To address these limitations, we identify two key challenges:
\begin{enumerate}
    \item Enhancing the search algorithm to include hardware architecture and parameters as adjustable factors.
    \item Refining the sampling process to better capture the solution distribution and minimize non-feasible configurations.
\end{enumerate}

We propose two main advancements in the co-optimizing compiler for DNNs. First, we integrate a multi-agent reinforcement learning (MARL) strategy into the search algorithm, enabling simultaneous optimization of hardware and software parameters. This approach leverages collective intelligence for a more holistic and efficient optimization. Second, we introduce a novel Confidence Sampling (CS) method to replace existing uniform \cite{chen2018tvm} and adaptive sampling methods \cite{ahn2020chameleon}, identifying configurations with a higher probability of success based on learned patterns.

Our primary optimization goal is to maximize throughput. The search space comprises configurations defined by various knobs, including tile dimensions for batch size, input channels, output channels, and threading options. By systematically exploring these configurations, our framework can adapt to different hardware platforms and optimize performance effectively.

\subsection{Overview of \alg}
\label{subsec:ovDCOC}

\begin{figure}
     \centering
     \includegraphics[width=0.45\textwidth]{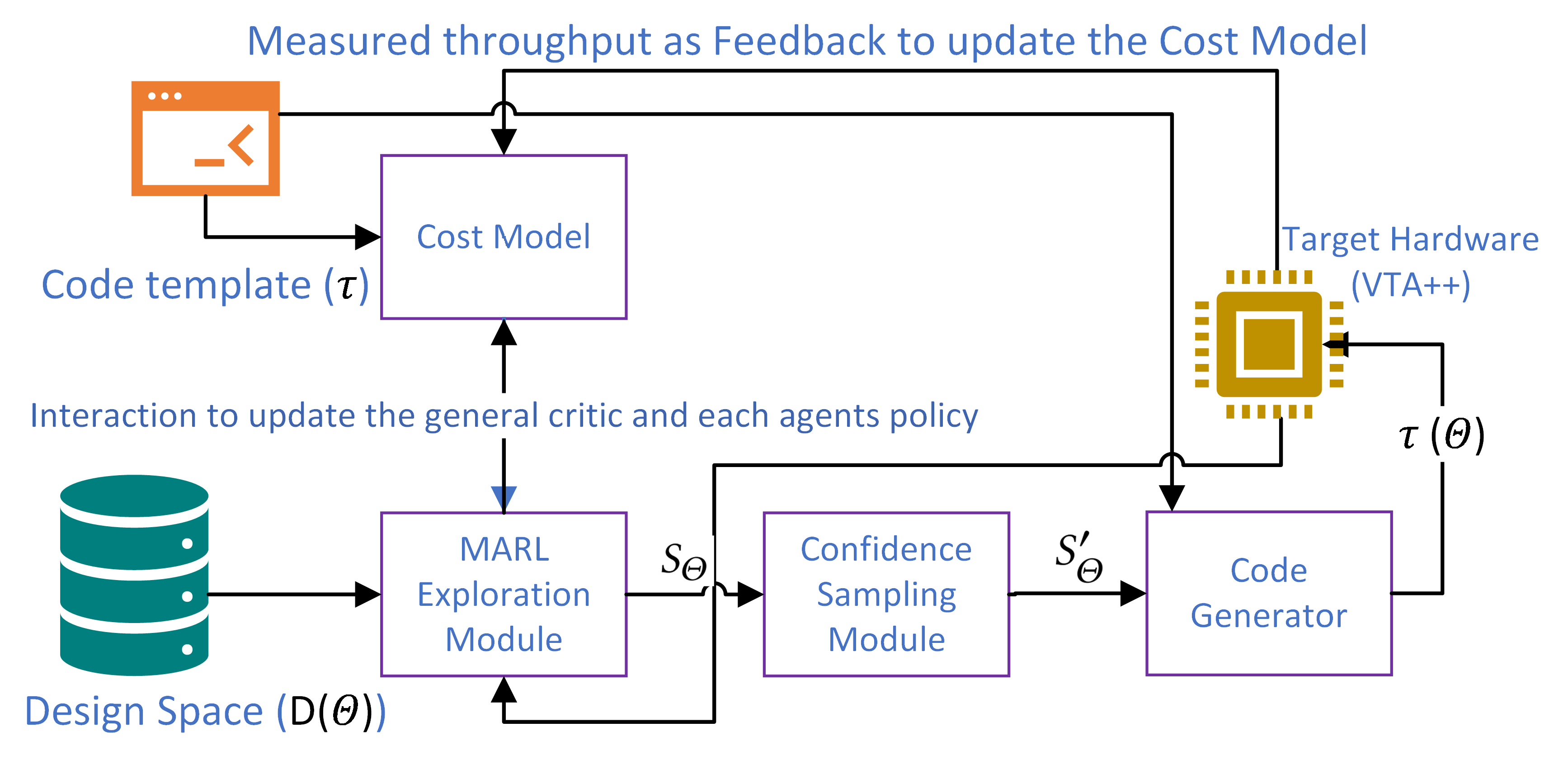}
     \vspace{-1em}
     \caption{\small Overall search flow of \alg.}
     \label{fig:DCOC}
\vspace{-2em}
\end{figure}

Figure ~\ref{fig:DCOC} depicts the architecture of our sophisticated co-optimizing compiler, termed \alg, which elucidates the process of optimization compilation. \alg initiates with a code template, $\tau$, for each layer of the neural network and a corresponding design space $D_{\Theta}$, delineating the range of possible configurations. The compiler then engages in an optimization process, navigating through the configuration space $\Theta$ to ascertain the optimal code configuration, represented as $\tau(\Theta^*)$.

\alg maneuvers through the design space, employing a cost model as a surrogate for direct hardware measurements. The MARL Exploration module generates an initial set of candidate configurations $S_{\Theta}$. Through the CS method, \alg refines $S_{\Theta}$ to produce a more focused subset $S'_{\Theta}$, which contains a condensed yet highly promising set of configurations. The configurations in $S'_{\Theta}$ are then passed to the MARL Code Generator module, which incorporates them with the input template $\tau$ to create a series of potential executable codes $\tau(\Theta)$. These are subsequently deployed on the hardware for empirical runtime evaluations. The hardware runtimes provide a measure of the configurations' fitness, captured by a fitness function $f$, which informs the update of the cost model and enhances the exploration in subsequent iterations. After several iterations, the process converges to some $\tau(\Theta^*)$ that achieves optimal fitness, characterized by the shortest hardware execution runtime. This configuration $(\Theta^*)$ is then selected as the output for that network layer.


\subsection{MARL Exploration}
\label{subsec:MARLEXP}

\begin{figure}[t]
     \centering
     \includegraphics[width=0.33\textwidth]{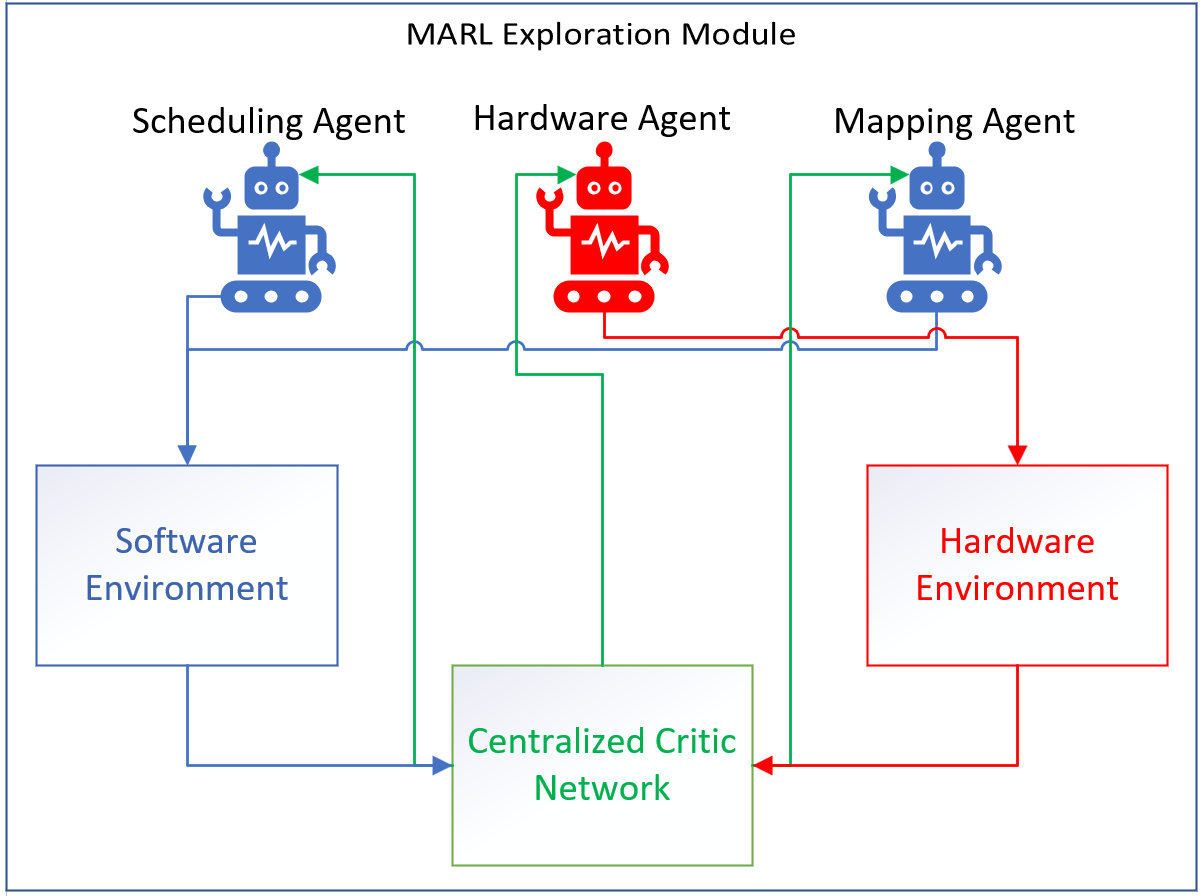}
     \vspace{-1.3em}
     \caption{\small High-level view of MARL Exploration Module. Each Agent has a policy network and, based on the centralized critic feedback, it will do an action in its own environment.}
     \label{fig:MARL}
\vspace{-2em}
\end{figure}

In our work, we implement a MARL Exploration module (As shown in Figure \ref{fig:MARL}) that employs a Multi-Agent Reinforcement Learning (MARL) strategy designed to optimize the configurations of DNN architectures and hardware simultaneously. This module utilizes the principle of Centralized Training and Decentralized Execution (CTDE), allowing agents to learn collaboratively while operating independently during the execution phase to maximize the fitness function, \(f\), of configurations within the configuration space, \(S_{\Theta}\). MARL's ability to manage complex decision-making environments makes it particularly suitable for navigating the high-dimensional configuration space of DNNs.

The MARL Exploration strategy employs three agents, each equipped with a policy network, as detailed in Table \ref{tab:agent_roles}. Additionally, there is a centralized critic network (value network) that, along with the policy networks, is implemented as a Multi-Layer Perceptron (MLP). The policy network directs each agent to propose adjustments to the configuration knobs (as shown in Table \ref{tab:vta_optimization}) within its assigned portion of the design space. In contrast, the value network estimates the potential value of these adjustments. A shared central critic aids in evaluating the global state, facilitating collective learning during the training phase, while each agent makes decisions independently during execution, adhering to the CTDE paradigm.

\begin{table}[t]
\centering
\caption{Roles and Responsibilities of Different Agents}
\vspace{-1em}
 \resizebox{\columnwidth}{!} {\begin{tabular}{|>{\raggedright\arraybackslash}m{1.6cm}|>{\raggedright\arraybackslash}m{6.4cm}|}
\hline
\textbf{Agent Type} & \textbf{Description} \\
\hline
\textbf{\begin{tabular}{@{}c@{}} Scheduling \\ Agent\end{tabular}} & Focuses on task parallelization and distribution for effective scheduling. \\
\hline
\textbf{\begin{tabular}{@{}c@{}}Mapping \\ Agent\end{tabular}} & Divides and processes data dimensions (tensor height and width) for optimal hardware computation mapping. \\
\hline
\textbf{\begin{tabular}{@{}c@{}}Hardware \\ Agent\end{tabular}} & Adjusts parameters for dividing and processing tensor components (batches, input and output channels) to optimize hardware utilization based on resource counts and performance levels. \\
\hline
\end{tabular}}

\label{tab:agent_roles}
\vspace{-1em}
\end{table}

The optimization unfolds through several episodes, each consisting of multiple search steps. During these steps, the agents evaluate the current configuration and independently decide on actions to improve subsequent configuration fitness, guided by shared and individual observations. The process detailed in Algorithm \ref{alg:ctdeMARL} outlines how each agent interacts within the CTDE, iterating through configurations to enhance the network performance.

\begin{algorithm}[htbp]
\caption{CTDE MARL Exploration for DNN Configuration Optimization}
\label{alg:ctdeMARL}
\begin{algorithmic}[1]
\State Initialize centralized critic and individual policy networks for each agent
\For{each episode}
    \State Initialize a set of configurations, \(S_{\Theta}\)
    \For{each search step in the episode}
        \For{each agent}
            \parState {%
            Observe the current configuration, \(\Theta\), and receive 
            
            shared insights from the centralized critic}
            \parState {%
            Independently chooses an action based on the policy 
            
            network and local observations}
            \State Apply the action to update the configuration
            \State Estimate the new configuration's local value
        \EndFor
        \parState {%
        Collectively evaluate all new configurations using the cost model}
        \parState {%
        Update the centralized critic based on global performance feedback}
        \parState {%
        Individually update each agent's policy network based on local and shared feedback}
    \EndFor
    \parState {%
    Determine the configuration with the highest fitness from collective insights}
\EndFor
\State Return the optimal configuration, \(\tau(\Theta^*)\)
\end{algorithmic}
\end{algorithm}

Each episode integrates the evaluations from a cost model that serves as a surrogate for direct hardware performance measurements. This model's feedback is used to update the centralized critic, enhance the shared knowledge base, and refine the individual policy networks, refining each agent's decision-making capabilities based on global and local feedback. As the episodes progress, the MARL Exploration Module becomes increasingly adept at identifying configurations that yield the best performance, ultimately converging on \(\tau(\Theta^*)\) with the optimal fitness, \(f\).

Applying this CTDE framework within the MARL context, coupled with the PPO optimization algorithm, represents a significant advancement in the auto-tuning of DNN and hardware configurations. This approach systematically improves the configuration search process, significantly enhancing DNN performance while reducing the computational overhead.

\subsubsection{Cost Model and Central Critic Update}
In the DCOC framework, the optimization goal is to improve throughput, with the cost model reflecting the inverse of execution time. The central critic, a key component of our MARL setup, updates its policy based on the aggregated feedback from all agents, continuously optimizing the global state evaluation. The critic's learning process is guided by the update rule, which is mathematically expressed in section \ref{subsec:ACMARL}, which minimizes the mean squared error between the predicted values and the actual rewards obtained.

\subsubsection{Incorporating Hardware and Software Constraints}
To integrate constraints related to hardware, such as area limitations, or software, such as memory usage, into the MARL framework, a penalty term can be incorporated into the reward function. This approach adjusts the reward based on the degree to which the constraints are violated, effectively guiding the agents to prefer configurations that adhere to these constraints. For instance, a penalty function $P$ can be defined as follows:

\begin{align}
P(\Theta) = & \lambda \left( \max(0, \text{area}(\Theta) - \text{area}_{\text{max}}) \right. \nonumber \\
& \left. + \max(0, \text{memory}(\Theta) - \text{memory}_{\text{max}}) \right)
\end{align}

\noindent where $\lambda$ is a scaling factor that adjusts the impact of the penalty, $\text{area}(\Theta)$ and $\text{memory}(\Theta)$ are the area and memory usage of the configuration $\Theta$, and $\text{area}_{\text{max}}$ and $\text{memory}_{\text{max}}$ are the respective maximums.

The reward function, modified to include this penalty, becomes:

\begin{equation}
R_t = \frac{1}{\text{execution time}(\Theta)} - P(\Theta)
\end{equation}

This modification ensures that the MARL agents not only seek to optimize performance but also adhere to specified design constraints, balancing efficiency with practical deployment considerations.
In this work, without loss of generality, we considered VTA++ \cite{banerjee2021highly} as the HW platform. The VTA++ architecture is comparable to other hardware platforms, such as GPUs and FPGAs, due to its configurable nature and similarity in processing elements \cite{banerjee2021highly}. This similarity ensures that the optimization strategies and results obtained from VTA++ can be generalized and applied to these other hardware platforms, validating the broad applicability of our framework.

\begin{table}[t]
\centering
\caption{Knobs in the design space to optimize convolution layers (These knobs make a search space as big as O(\(2^{12}\)))}
\vspace{-1.2em}
\resizebox{\columnwidth}{!}{\begin{tabular}{|m{3.5cm}|m{5.5cm}|}
\hline
\textbf{Agent Type} & \textbf{Knobs} \\
\hline
\textbf{\begin{tabular}{@{}c@{}}Hardware Design Agent \\ (Hardware Optimizer)\end{tabular}} & - Tile across batch size (tile\_b) \newline
- Tile across input channels (tile\_ci) \newline
- Tile across output channels (tile\_co) \\
\hline
\textbf{\begin{tabular}{@{}c@{}}Scheduling Agent \\ (Software Optimizer)\end{tabular}} & - Horizontal threading (h\_threading) \newline
- Output channel threading (oc\_threading) \\
\hline
\textbf{\begin{tabular}{@{}c@{}}Mapping Agent \\ (Software Optimizer)\end{tabular}} & - Tile across height (tile\_h) \newline
- Tile across width (tile\_w) \\
\hline
\end{tabular}}

\label{tab:vta_optimization}
\vspace{-1.8em}
\end{table}

\subsection{Confidence Sampling}
\label{subsec:confsamp}

In DNN optimization domain, the ability to efficiently sample configuration spaces without costly hardware measurements is essential for rapid compilation. We introduce the Confidence Sampling (CS) method. This technique is designed to leverage the probability distribution of configurations' fitness judiciously to select a subset likely to yield high-performing network configurations.

Although the Confidence Sampling method operates on the principles of probability-guided selection and shares some conceptual similarities with Importance Sampling (IS) \cite{tokdar2010importance} in statistical methods, it is distinct in its purpose and application. In IS, samples are re-weighted based on their probability density to approximate expected values more efficiently. Confidence Sampling, however, specifically focuses on reducing the number of hardware measurements required when exploring the design space. By evaluating the configurations' values, this method synthesizes a subset that not only encapsulates the diversity of the configuration space but also emphasizes the high-confidence regions, i.e., areas where the likelihood of encountering an (near-) optimal configuration is high.

The core of the CS method lies in its algorithmic approach to filtering and enhancing the set of configurations for subsequent optimization iterations. To formalize this process, we introduce Algorithm~\ref{alg:confident_sampling}, which captures the essence of the CS method within the optimization pipeline. The process involves the following steps:

\begin{enumerate}
    \item \textbf{Evaluate Configurations:} Each configuration's value is estimated using the output of the value network (critic network) to construct a probability distribution over the configuration space. (Line 2)
    \item \textbf{Probability-Guided Selection:} Configurations are sampled based on this distribution, prioritizing those with higher estimated values. (Line 3-4)
    \item \textbf{Confidence Assessment:} A dynamic thresholding mechanism is applied to determine the confidence level of each selected configuration. (Line 5)
    \item \textbf{Synthesis of Configurations:} Lower-confidence configurations are replaced with synthesized ones by combining each parameter's most frequently occurring settings across the sampled configurations. (Line 6-7)
\end{enumerate}

\begin{algorithm}
\caption{Confidence Sampling for DNN Optimization}
\label{alg:confident_sampling}
\begin{algorithmic}[1]
\Procedure{ConfidenceSampling}{$\mathcal{S}_{\Theta}$, $value\_network$, $N_{select}$}
    \parState {%
    $\mathcal{V}_{preds} \gets value\_network.predict(\mathcal{S}_{\Theta})$ \Comment{Estimate values for all configurations}}
    \parState {%
    $\mathcal{P} \gets softmax(\mathcal{V}_{preds})$ \Comment{Convert values to a probability distribution}}
    \State $\mathcal{C}_{selected} \gets \Call{SelectConfigurations}{\mathcal{P}, N_{configs}}$
    \State $threshold \gets \Call{ComputeDynamicThreshold}{\mathcal{V}_{preds}}$ \Comment{Compute an adaptive threshold}
    \State $\mathcal{C}_{high\_conf} \gets \{ c \in \mathcal{C}_{selected} \mid \mathcal{V}_{preds}[c] > threshold \}$
    \State \textbf{return} $\mathcal{C}_{high\_conf}$
\EndProcedure

\Function{SelectConfigurations}{$\mathcal{P}$, $N_{configs}$}
     \parState {%
     $indices \gets$ sample indices from $[1, \ldots, |\mathcal{P}|]$ with probability $\mathcal{P}$ }
    \State \textbf{return} $\mathcal{S}_{\Theta}[indices[1:N_{configs}]]$
\EndFunction

\Function{ComputeDynamicThreshold}{$\mathcal{V}_{preds}$}
    \State \textbf{return} $median(\mathcal{V}_{preds})$ 
\EndFunction

\end{algorithmic}
\vspace{-0.3em}
\end{algorithm}

\begin{figure}[t]
     \centering
     \begin{subfigure}[b]{0.45\textwidth}
         \centering
         \includegraphics[width=\textwidth]{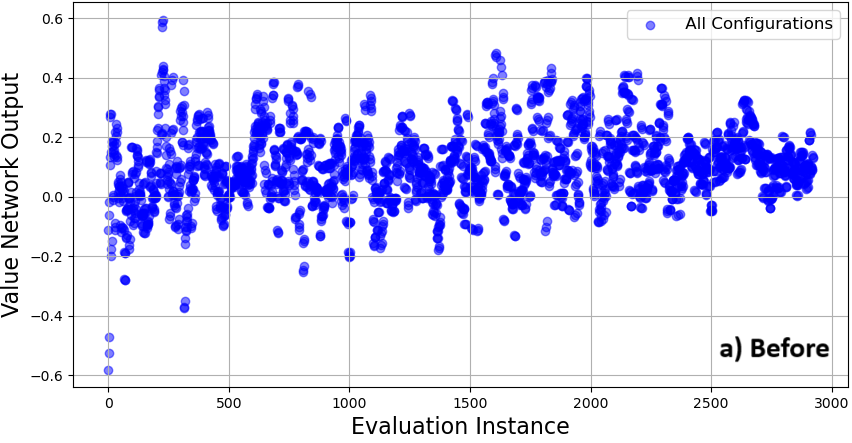}
     \end{subfigure}
     \vspace{-0.3em}
     \vfill
     \begin{subfigure}[b]{0.45\textwidth}
         \centering
         \includegraphics[width=\textwidth]{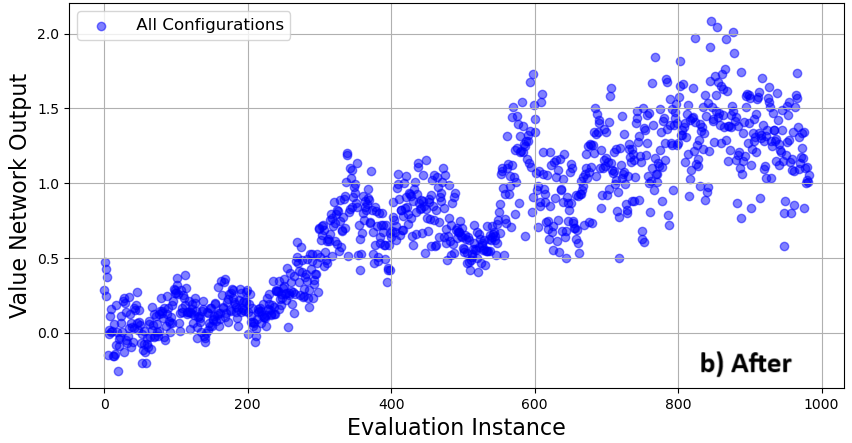}
     \end{subfigure}
     \vspace{-1.5em}
    \caption{Configurations over time for ResNet-18 model a) before and b) after applying the CS method.}
    \label{fig:conf}
\vspace{-2em}    
\end{figure}

By applying the CS method, we have observed a reduced number of required configurations, as shown in Figure ~\ref{fig:conf}. Moreover, we can observe the sampling process gravitates towards configurations that demonstrate superior performance over time, highlighting the efficacy of this method.

\section{Experimental Results}
\label{sec:exp-res}

\subsection{Experimental Setup}
\label{subsec:exp-set}

We have incorporated \alg within the TVM framework to thoroughly evaluate its components, contrasting its performance with that of AutoTVM and CHAMELEON. This integration allows for a comprehensive, end-to-end assessment of \alg across a suite of advanced deep-learning models, including AlexNet, VGG-11, VGG-13, VGG-16, VGG-19, ResNet-18, and ResNet-34. All models were benchmarked using the same AutoTVM compilation duration to ensure a rigorous and equitable comparison. Further, we have employed the source code of the CHAMELEON \cite{ahn2020chameleon} to compare its efficacy with AutoTVM and \alg. 
As mentioned above, AutoTVM and CHAMELEON do not support hardware configuration exploration. Hence, we also added the hardware architecture agent for them to have a fair comparison.
Performance tests were conducted on a robust platform powered by a 3.4 GHz AMD EPYC 7763 64-Core Processor, providing a stable and powerful environment for these evaluations. As discussed before, we used the VTA++ simulator as our target hardware. 

In our setup, each agent is equipped with its own policy network, and a centralized value network is the critic for all agents. Both networks are implemented as Multi-Layer Perceptrons (MLPs) using TensorFlow \cite{abadi2016tensorflow} that their details are:

\begin{itemize}
\item \textit{Policy Network:} Each agent’s policy network consists of a single hidden layer with 20 neurons using ReLU activation functions. The output layer uses softmax activation to generate a probability distribution over actions, allowing for decision-making informed by learned policies.

\item \textit{Value Network:} The centralized critic, or value network, employs three hidden layers with 20 neurons each, using tanh activation functions. This design helps stabilize learning by providing a continuous estimate of state values, which guides collective policy adjustments.
\end{itemize}

Hyper-parameter tuning is crucial in optimizing the performance of machine learning tools and models \cite{yu2020hyper}. To facilitate transparency and reproducibility, we have detailed the hyper-parameters utilized in our evaluation in Table \ref{tab:DCOC_hyperparameters}. These hyper-parameters were meticulously tuned offline to optimize model performance, and they are the same as those used in CHAMELEON \cite{ahn2020chameleon}. For the hyper-parameters listed in Table \ref{tab:hyperparameters}, we adhered to the values specified in the AutoTVM study to maintain consistency and ensure a fair comparison. Similarly, we adopted the hyper-parameter settings from the MAPPO paper \cite{yu2022surprising} for the MARL exploration module, aligning our methodologies with established research to validate our findings effectively. 

\begin{table}[h]
\vspace{-0.8em}
\centering
\caption{Hyper-parameters (HPs) used in \alg.}
\label{tab:DCOC_hyperparameters}
\vspace{-1em}
\small
\begin{tabular}{@{}lll@{}}
\toprule
\textbf{HP} & \textbf{Value} & \textbf{Description} \\ \midrule
\textit{iteration\_opt} & 16 & \begin{tabular}{@{}l@{}} Total iterations for the optimization cycle \\ (equivalent to 1000 hardware measurements) \end{tabular} \\
\textit{modeGBT} & xgb-reg & Loss function type utilized in the cost model \\
\textit{bGBT} & 64 & Planning's maximum batch size in GBT \cite{chen2018tvm} \\
\textit{episode\_rl} & 128 & Number of episodes for the RL process \\
\textit{step\_rl} & 500 & Maximum \# of steps in a single RL episode \\ \bottomrule
\end{tabular}
\vspace{-1em}
\end{table}

\begin{table}[h]
\centering
\caption{Hyper-parameters (HPs) used in AutoTVM}
\label{tab:hyperparameters}
\vspace{-1em}
\small
\begin{threeparttable}
\begin{tabular}{@{}lll@{}}
\toprule
\textbf{HP} & \textbf{Value} & \textbf{Description} \\ \midrule
$\Sigma(b_{GBT})$ & 1000 & Total count of hardware measurements \\
$mode_{GBT}$ & xgb-reg & Loss function type for the cost model \\
$b_{GBT}$ & 64 & Batch size for planning in GBT \cite{chen2018tvm} \\
$n_{sa}$ & 128 & Count of Markov chains in parallel SA$^*$ \\
$step_{sa}$ & 500 & Maximum \# of steps in a single SA run \\ \bottomrule
\end{tabular}
\begin{tablenotes}
  \footnotesize
  \item $^*$SA: Simulated Annealing.
\end{tablenotes}
\end{threeparttable}
\vspace{-2em}
\end{table}

\subsection{End-to-end Evaluation}

\begin{figure}[h]
     \centering
     \includegraphics[width=0.45\textwidth]{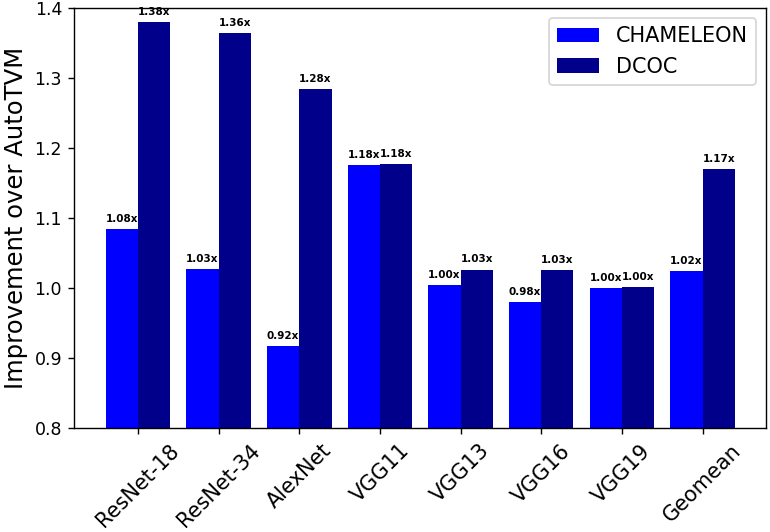}
     \caption{\small Comparing the achieved throughput of different frameworks over AutoTVM on VTA++.}
     \label{fig:Throughput}
\vspace{-1em}     
\end{figure}

Figure \ref{fig:Throughput} Compares the achieved throughput of different frameworks over AutoTVM. As the results illustrate, the significant reduction in mean inference time (advancement in throughput) achieved by \alg compared to AutoTVM and CHAMELEON for various DNN models: AlexNet, VGG-11, VGG-13, VGG-16, VGG-19, and ResNet-18, and ResNet-34. On average, our methodology demonstrates a 1.17× improvement in throughput. Table \ref{tab:mean_inference_times} reports detailed numerical results supporting these findings.

\begin{table}[htbp]
\centering
\caption{Mean inference times for different frameworks on VTA++ (in seconds)}
\label{tab:mean_inference_times}
\begin{tabular}{@{}lSSS@{}}
\toprule
Model    & {AutoTVM} & {CHAMELEON} & {DCOC} \\ \midrule
ResNet-18 & 1.73061  & 1.59595     & \textbf{1.25448} \\
ResNet-34 & 3.63409  & 3.53780     & \textbf{2.66561} \\
AlexNet   & 0.54116  & 0.58981     & \textbf{0.42160}  \\
VGG11     & 4.95229  & 4.21283     & \textbf{4.20685} \\
VGG13     & 6.19289  & 6.16350     & \textbf{6.03418} \\
VGG16     & 8.55878  & 8.73089     & \textbf{8.34791} \\
VGG19     & 11.29150 & 11.28839    & \textbf{11.27800} \\ \bottomrule
\end{tabular}
\vspace{-1.2em}
\end{table}

\begin{figure}[h]
     \centering
     \includegraphics[width=0.45\textwidth]{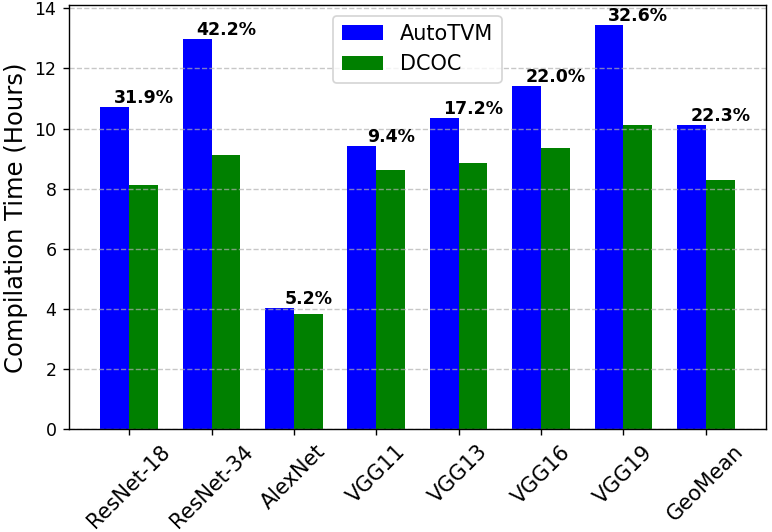}
     \caption{\small Comparing the compilation time of different frameworks (The percentages show the speedup of DCOC compared to AutoTVM).}
     \label{fig:compile}
\vspace{-1em}       
\end{figure}

\begin{figure}[t]
     \centering
     \includegraphics[width=0.45\textwidth]{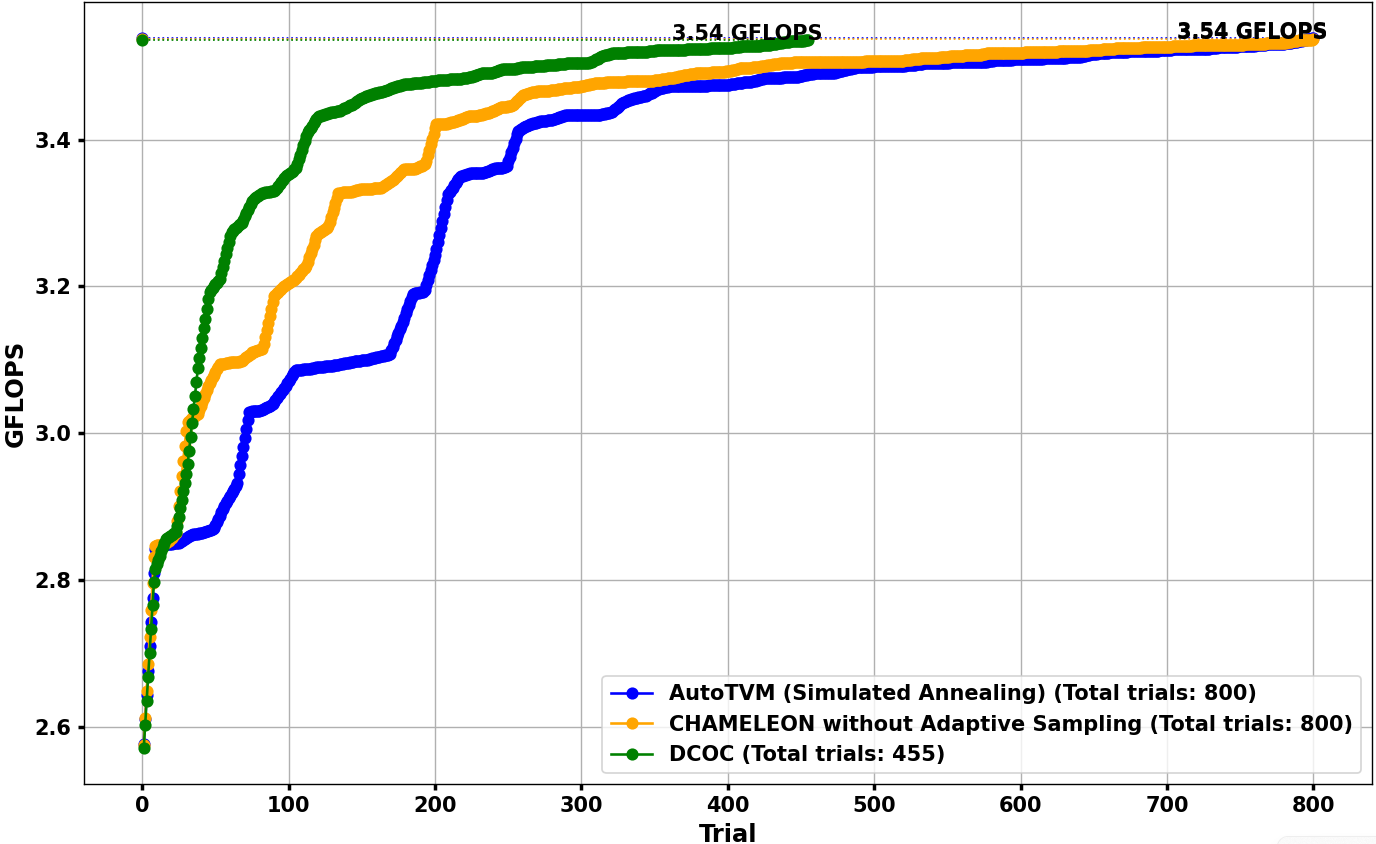}   
     \vspace{-1.4em}
     \caption{\small Comparing the compiled code performance (GFLOPS per Trial) of different frameworks for ResNet-18 model.}
     \label{fig:GFLOPS}
\end{figure}

While achieving improvement in throughput, DCOC also needs less compilation time compared to AutoTVM; Figure \ref{fig:compile} shows that DCOC speeds up the optimization time (the entire process of compiling) up to 42.2\%. 
Figure \ref{fig:GFLOPS} shows DCOC's comparative output code performance relative to other frameworks. Notably, DCOC achieves convergence to the peak GFLOPS value, equivalent to those reached by AutoTVM and CHAMELEON (without adaptive sampling), but with greater efficiency. This is accomplished with fewer hardware measurements and at a faster rate, underscoring the efficacy of the CS method.

\section{Conclusion}
\label{sec:conclusion}

This paper introduced DCOC, an innovative multi-agent reinforcement learning-based compiler that significantly advances the field of DNN accelerator design. By integrating a multi-agent system with specialized roles—two agents focused on software optimization and one on hardware—DCOC addressed the intricacies of DNN architecture and effectively navigated the vast configuration space to enhance performance and efficiency. Employing the MARL Exploration module and Confidence Sampling method, DCOC reduced the computational overhead and expedited the optimization process while increasing the throughput.




\bibliographystyle{ACM-Reference-Format}
\bibliography{sample-base}

\end{document}

%% file: preamble.tex
\copyrightyear{2025}
\acmYear{2025}
\setcopyright{rightsretained}
\acmConference[ASPDAC '25]{30th Asia and South Pacific Design Automation Conference}{January 20--23, 2025}{Tokyo, Japan}
\acmBooktitle{30th Asia and South Pacific Design Automation Conference (ASPDAC '25), January 20--23, 2025, Tokyo, Japan}\acmDOI{10.1145/3658617.3697547}
\acmISBN{979-8-4007-0635-6/25/01}

\usepackage{amsthm} 

\usepackage{mathtools}

\usepackage{fixmath}

\usepackage{multirow}
\usepackage{tabularx}
\newcolumntype{L}{>{\RaggedRight\arraybackslash}X}
\newcolumntype{C}{>{\centering\arraybackslash}X}
\usepackage{dcolumn}
\usepackage{colortbl}
\usepackage{siunitx}

\usepackage{algorithm}
\usepackage{algpseudocode}
\algnewcommand\parState[1]{\State\parbox[t]{\dimexpr\linewidth-\algorithmicindent}{#1\strut}}

\usepackage{url}

\usepackage{subcaption}

%% file: main.bbl

\begin{thebibliography}{33}


\ifx \showCODEN    \undefined \def \showCODEN     #1{\unskip}     \fi
\ifx \showDOI      \undefined \def \showDOI       #1{#1}\fi
\ifx \showISBNx    \undefined \def \showISBNx     #1{\unskip}     \fi
\ifx \showISBNxiii \undefined \def \showISBNxiii  #1{\unskip}     \fi
\ifx \showISSN     \undefined \def \showISSN      #1{\unskip}     \fi
\ifx \showLCCN     \undefined \def \showLCCN      #1{\unskip}     \fi
\ifx \shownote     \undefined \def \shownote      #1{#1}          \fi
\ifx \showarticletitle \undefined \def \showarticletitle #1{#1}   \fi
\ifx \showURL      \undefined \def \showURL       {\relax}        \fi
\providecommand\bibfield[2]{#2}
\providecommand\bibinfo[2]{#2}
\providecommand\natexlab[1]{#1}
\providecommand\showeprint[2][]{arXiv:#2}

\bibitem[Abadi et~al\mbox{.}(2016)]%
        {abadi2016tensorflow}
\bibfield{author}{\bibinfo{person}{Mart{\'\i}n Abadi}, \bibinfo{person}{Paul Barham}, \bibinfo{person}{Jianmin Chen}, \bibinfo{person}{Zhifeng Chen}, \bibinfo{person}{Andy Davis}, \bibinfo{person}{Jeffrey Dean}, \bibinfo{person}{Matthieu Devin}, \bibinfo{person}{Sanjay Ghemawat}, \bibinfo{person}{Geoffrey Irving}, \bibinfo{person}{Michael Isard}, {et~al\mbox{.}}} \bibinfo{year}{2016}\natexlab{}.
\newblock \showarticletitle{$\{$TensorFlow$\}$: a system for $\{$Large-Scale$\}$ machine learning}. In \bibinfo{booktitle}{\emph{12th USENIX symposium on operating systems design and implementation (OSDI 16)}}. \bibinfo{pages}{265--283}.
\newblock


\bibitem[Ahn et~al\mbox{.}(2022)]%
        {ahn2022glimpse}
\bibfield{author}{\bibinfo{person}{Byung~Hoon Ahn}, \bibinfo{person}{Sean Kinzer}, {and} \bibinfo{person}{Hadi Esmaeilzadeh}.} \bibinfo{year}{2022}\natexlab{}.
\newblock \showarticletitle{Glimpse: mathematical embedding of hardware specification for neural compilation}. In \bibinfo{booktitle}{\emph{Proceedings of the 59th ACM/IEEE Design Automation Conference}}. \bibinfo{pages}{1165--1170}.
\newblock


\bibitem[Ahn et~al\mbox{.}(2020)]%
        {ahn2020chameleon}
\bibfield{author}{\bibinfo{person}{Byung~Hoon Ahn}, \bibinfo{person}{Prannoy Pilligundla}, \bibinfo{person}{Amir Yazdanbakhsh}, {and} \bibinfo{person}{Hadi Esmaeilzadeh}.} \bibinfo{year}{2020}\natexlab{}.
\newblock \showarticletitle{Chameleon: Adaptive code optimization for expedited deep neural network compilation}.
\newblock \bibinfo{journal}{\emph{arXiv preprint arXiv:2001.08743}} (\bibinfo{year}{2020}).
\newblock


\bibitem[Allen and Kennedy(2002)]%
        {allen2002optimizing}
\bibfield{author}{\bibinfo{person}{Randy Allen} {and} \bibinfo{person}{Ken Kennedy}.} \bibinfo{year}{2002}\natexlab{}.
\newblock \showarticletitle{Optimizing compilers for modern architectures: a dependence-baced approach}.
\newblock \bibinfo{journal}{\emph{(No Title)}} (\bibinfo{year}{2002}).
\newblock


\bibitem[Ashrafi et~al\mbox{.}(2024)]%
        {Ashrafi2024.06.26.24309553}
\bibfield{author}{\bibinfo{person}{Negin Ashrafi}, \bibinfo{person}{Armin Abdollahi}, \bibinfo{person}{Greg Placencia}, {and} \bibinfo{person}{Maryam Pishgar}.} \bibinfo{year}{2024}\natexlab{}.
\newblock \showarticletitle{Process Mining/Deep Learning Model to Predict Mortality in Coronary Artery Disease Patients}.
\newblock \bibinfo{journal}{\emph{medRxiv}} (\bibinfo{year}{2024}).
\newblock
\urldef\tempurl%
\url{https://doi.org/10.1101/2024.06.26.24309553}
\showDOI{\tempurl}
\showeprint{https://www.medrxiv.org/content/early/2024/06/27/2024.06.26.24309553.full.pdf}


\bibitem[Azizi et~al\mbox{.}(2024)]%
        {azizi2024efficientnoisemitigationenhancing}
\bibfield{author}{\bibinfo{person}{Seyedarmin Azizi}, \bibinfo{person}{Mohammad~Erfan Sadeghi}, \bibinfo{person}{Mehdi Kamal}, {and} \bibinfo{person}{Massoud Pedram}.} \bibinfo{year}{2024}\natexlab{}.
\newblock \bibinfo{title}{Efficient Noise Mitigation for Enhancing Inference Accuracy in DNNs on Mixed-Signal Accelerators}.
\newblock
\newblock
\showeprint[arxiv]{2409.18553}~[cs.LG]
\urldef\tempurl%
\url{https://arxiv.org/abs/2409.18553}
\showURL{%
\tempurl}


\bibitem[Bakshi and Johnsson(2023)]%
        {bakshi2023computationally}
\bibfield{author}{\bibinfo{person}{Suyash Bakshi} {and} \bibinfo{person}{Lennart Johnsson}.} \bibinfo{year}{2023}\natexlab{}.
\newblock \showarticletitle{Computationally Efficient DNN Mapping Search Heuristic using Deep Reinforcement Learning}.
\newblock \bibinfo{journal}{\emph{ACM Transactions on Embedded Computing Systems}} \bibinfo{volume}{22}, \bibinfo{number}{5s} (\bibinfo{year}{2023}), \bibinfo{pages}{1--21}.
\newblock


\bibitem[Banerjee et~al\mbox{.}(2021)]%
        {banerjee2021highly}
\bibfield{author}{\bibinfo{person}{Suvadeep Banerjee}, \bibinfo{person}{Steve Burns}, \bibinfo{person}{Pasquale Cocchini}, \bibinfo{person}{Abhijit Davare}, \bibinfo{person}{Shweta Jain}, \bibinfo{person}{Desmond Kirkpatrick}, \bibinfo{person}{Anton Sorokin}, \bibinfo{person}{Jin Yang}, {and} \bibinfo{person}{Zhenkun Yang}.} \bibinfo{year}{2021}\natexlab{}.
\newblock \showarticletitle{A highly configurable hardware/Software stack for DNN inference acceleration}.
\newblock \bibinfo{journal}{\emph{arXiv preprint arXiv:2111.15024}} (\bibinfo{year}{2021}).
\newblock


\bibitem[Chen and Guestrin(2016)]%
        {chen2016xgboost}
\bibfield{author}{\bibinfo{person}{Tianqi Chen} {and} \bibinfo{person}{Carlos Guestrin}.} \bibinfo{year}{2016}\natexlab{}.
\newblock \showarticletitle{Xgboost: A scalable tree boosting system}. In \bibinfo{booktitle}{\emph{Proceedings of the 22nd acm sigkdd international conference on knowledge discovery and data mining}}. \bibinfo{pages}{785--794}.
\newblock


\bibitem[Chen et~al\mbox{.}(2018)]%
        {chen2018tvm}
\bibfield{author}{\bibinfo{person}{Tianqi Chen}, \bibinfo{person}{Thierry Moreau}, \bibinfo{person}{Ziheng Jiang}, \bibinfo{person}{Lianmin Zheng}, \bibinfo{person}{Eddie Yan}, \bibinfo{person}{Haichen Shen}, \bibinfo{person}{Meghan Cowan}, \bibinfo{person}{Leyuan Wang}, \bibinfo{person}{Yuwei Hu}, \bibinfo{person}{Luis Ceze}, {et~al\mbox{.}}} \bibinfo{year}{2018}\natexlab{}.
\newblock \showarticletitle{$\{$TVM$\}$: An automated $\{$End-to-End$\}$ optimizing compiler for deep learning}. In \bibinfo{booktitle}{\emph{13th USENIX Symposium on Operating Systems Design and Implementation (OSDI 18)}}. \bibinfo{pages}{578--594}.
\newblock


\bibitem[Dhakal et~al\mbox{.}(2022)]%
        {dhakal2022slice}
\bibfield{author}{\bibinfo{person}{Aditya Dhakal}, \bibinfo{person}{KK Ramakrishnan}, \bibinfo{person}{Sameer~G Kulkarni}, \bibinfo{person}{Puneet Sharma}, {and} \bibinfo{person}{Junguk Cho}.} \bibinfo{year}{2022}\natexlab{}.
\newblock \showarticletitle{Slice-tune: A system for high performance dnn autotuning}. In \bibinfo{booktitle}{\emph{Proceedings of the 23rd ACM/IFIP International Middleware Conference}}. \bibinfo{pages}{228--240}.
\newblock


\bibitem[Du and Ding(2021)]%
        {du2021survey}
\bibfield{author}{\bibinfo{person}{Wei Du} {and} \bibinfo{person}{Shifei Ding}.} \bibinfo{year}{2021}\natexlab{}.
\newblock \showarticletitle{A survey on multi-agent deep reinforcement learning: from the perspective of challenges and applications}.
\newblock \bibinfo{journal}{\emph{Artificial Intelligence Review}} \bibinfo{volume}{54}, \bibinfo{number}{5} (\bibinfo{year}{2021}), \bibinfo{pages}{3215--3238}.
\newblock


\bibitem[Farsi et~al\mbox{.}(2024)]%
        {202410.1684}
\bibfield{author}{\bibinfo{person}{Farhan Farsi}, \bibinfo{person}{Sadra Sabouri}, \bibinfo{person}{Kian Kashfipour}, \bibinfo{person}{Soroush Gooran}, \bibinfo{person}{Hossein Sameti}, {and} \bibinfo{person}{Ehsaneddin Asgari}.} \bibinfo{year}{2024}\natexlab{}.
\newblock \showarticletitle{SynTran-fa: Generating Comprehensive Answers for Farsi QA Pairs via Syntactic Transformation}.
\newblock \bibinfo{journal}{\emph{Preprints}} (\bibinfo{date}{October} \bibinfo{year}{2024}).
\newblock
\urldef\tempurl%
\url{https://doi.org/10.20944/preprints202410.1684.v1}
\showDOI{\tempurl}


\bibitem[Fayyazi et~al\mbox{.}(2024)]%
        {10.1145/3649476.3658736}
\bibfield{author}{\bibinfo{person}{Arash Fayyazi}, \bibinfo{person}{Mahdi Nazemi}, \bibinfo{person}{Arya Fayyazi}, {and} \bibinfo{person}{Massoud Pedram}.} \bibinfo{year}{2024}\natexlab{}.
\newblock \showarticletitle{NeuroBlend: Towards Low-Power yet Accurate Neural Network-Based Inference Engine Blending Binary and Fixed-Point Convolutions}. In \bibinfo{booktitle}{\emph{Proceedings of the Great Lakes Symposium on VLSI 2024}} (Clearwater, FL, USA) \emph{(\bibinfo{series}{GLSVLSI '24})}. \bibinfo{publisher}{Association for Computing Machinery}, \bibinfo{address}{New York, NY, USA}, \bibinfo{pages}{730–735}.
\newblock
\showISBNx{9798400706059}
\urldef\tempurl%
\url{https://doi.org/10.1145/3649476.3658736}
\showDOI{\tempurl}


\bibitem[Hossain et~al\mbox{.}(2023)]%
        {hossain2023computational}
\bibfield{author}{\bibinfo{person}{Md~Bipul Hossain}, \bibinfo{person}{Na Gong}, {and} \bibinfo{person}{Mohamed Shaban}.} \bibinfo{year}{2023}\natexlab{}.
\newblock \showarticletitle{Computational Complexity Reduction Techniques for Deep Neural Networks: A Survey}. In \bibinfo{booktitle}{\emph{2023 IEEE International Conference on Artificial Intelligence, Blockchain, and Internet of Things (AIBThings)}}. IEEE, \bibinfo{pages}{1--6}.
\newblock


\bibitem[Kumar et~al\mbox{.}(2021)]%
        {kumar2021data}
\bibfield{author}{\bibinfo{person}{Aviral Kumar}, \bibinfo{person}{Amir Yazdanbakhsh}, \bibinfo{person}{Milad Hashemi}, \bibinfo{person}{Kevin Swersky}, {and} \bibinfo{person}{Sergey Levine}.} \bibinfo{year}{2021}\natexlab{}.
\newblock \showarticletitle{Data-driven offline optimization for architecting hardware accelerators}.
\newblock \bibinfo{journal}{\emph{arXiv preprint arXiv:2110.11346}} (\bibinfo{year}{2021}).
\newblock


\bibitem[Lu et~al\mbox{.}(2022)]%
        {lu2022semi}
\bibfield{author}{\bibinfo{person}{Bingqian Lu}, \bibinfo{person}{Zheyu Yan}, \bibinfo{person}{Yiyu Shi}, {and} \bibinfo{person}{Shaolei Ren}.} \bibinfo{year}{2022}\natexlab{}.
\newblock \showarticletitle{A Semi-Decoupled Approach to Fast and Optimal Hardware-Software Co-Design of Neural Accelerators}.
\newblock \bibinfo{journal}{\emph{arXiv preprint arXiv:2203.13921}} (\bibinfo{year}{2022}).
\newblock


\bibitem[Lyu et~al\mbox{.}(2021)]%
        {lyu2021contrasting}
\bibfield{author}{\bibinfo{person}{Xueguang Lyu}, \bibinfo{person}{Yuchen Xiao}, \bibinfo{person}{Brett Daley}, {and} \bibinfo{person}{Christopher Amato}.} \bibinfo{year}{2021}\natexlab{}.
\newblock \showarticletitle{Contrasting centralized and decentralized critics in multi-agent reinforcement learning}.
\newblock \bibinfo{journal}{\emph{arXiv preprint arXiv:2102.04402}} (\bibinfo{year}{2021}).
\newblock


\bibitem[Mokhtari~Dowlatabad et~al\mbox{.}(2023)]%
        {diagnostics13020179}
\bibfield{author}{\bibinfo{person}{Hadi Mokhtari~Dowlatabad}, \bibinfo{person}{Amir Mamdouh}, \bibinfo{person}{Narges Yousefpour}, \bibinfo{person}{Reihane Mahdavi}, \bibinfo{person}{Ashkan Zandi}, \bibinfo{person}{Parisa Hoseinpour}, \bibinfo{person}{Seyed Mohammad~Sadegh Moosavi-Kiasari}, \bibinfo{person}{Fereshte Abbasvandi}, \bibinfo{person}{Yasin Kordehlachin}, \bibinfo{person}{Mohammad Parniani}, \bibinfo{person}{Karim Mohammadpour-Aghdam}, \bibinfo{person}{Pooya Faranoush}, \bibinfo{person}{Mohammad~Reza Foroughi-Gilvaee}, {and} \bibinfo{person}{Mohammad Abdolahad}.} \bibinfo{year}{2023}\natexlab{}.
\newblock \showarticletitle{High-Frequency (30 MHz–6 GHz) Breast Tissue Characterization Stabilized by Suction Force for Intraoperative Tumor Margin Assessment}.
\newblock \bibinfo{journal}{\emph{Diagnostics}} \bibinfo{volume}{13}, \bibinfo{number}{2} (\bibinfo{year}{2023}).
\newblock
\showISSN{2075-4418}
\urldef\tempurl%
\url{https://doi.org/10.3390/diagnostics13020179}
\showDOI{\tempurl}


\bibitem[Paszke et~al\mbox{.}(2019)]%
        {paszke2019pytorch}
\bibfield{author}{\bibinfo{person}{Adam Paszke}, \bibinfo{person}{Sam Gross}, \bibinfo{person}{Francisco Massa}, \bibinfo{person}{Adam Lerer}, \bibinfo{person}{James Bradbury}, \bibinfo{person}{Gregory Chanan}, \bibinfo{person}{Trevor Killeen}, \bibinfo{person}{Zeming Lin}, \bibinfo{person}{Natalia Gimelshein}, \bibinfo{person}{Luca Antiga}, {et~al\mbox{.}}} \bibinfo{year}{2019}\natexlab{}.
\newblock \showarticletitle{Pytorch: An imperative style, high-performance deep learning library}.
\newblock \bibinfo{journal}{\emph{Advances in neural information processing systems}}  \bibinfo{volume}{32} (\bibinfo{year}{2019}).
\newblock


\bibitem[Razmara et~al\mbox{.}(2024)]%
        {razmara2024feverdetectioninfraredthermography}
\bibfield{author}{\bibinfo{person}{Parsa Razmara}, \bibinfo{person}{Tina Khezresmaeilzadeh}, {and} \bibinfo{person}{B.~Keith Jenkins}.} \bibinfo{year}{2024}\natexlab{}.
\newblock \bibinfo{title}{Fever Detection with Infrared Thermography: Enhancing Accuracy through Machine Learning Techniques}.
\newblock
\newblock
\showeprint[arxiv]{2407.15302}~[cs.LG]
\urldef\tempurl%
\url{https://arxiv.org/abs/2407.15302}
\showURL{%
\tempurl}


\bibitem[Rieber et~al\mbox{.}(2022)]%
        {rieber2022hw}
\bibfield{author}{\bibinfo{person}{Dennis Rieber}, \bibinfo{person}{Moritz Reiber}, \bibinfo{person}{Oliver Bringmann}, {and} \bibinfo{person}{Holger Fr{\"o}ning}.} \bibinfo{year}{2022}\natexlab{}.
\newblock \showarticletitle{Hw-aware initialization of dnn auto-tuning to improve exploration time and robustness}.
\newblock \bibinfo{journal}{\emph{arXiv preprint arXiv:2205.15568}} (\bibinfo{year}{2022}).
\newblock


\bibitem[Ryu et~al\mbox{.}(2022)]%
        {ryu2022one}
\bibfield{author}{\bibinfo{person}{Jaehun Ryu}, \bibinfo{person}{Eunhyeok Park}, {and} \bibinfo{person}{Hyojin Sung}.} \bibinfo{year}{2022}\natexlab{}.
\newblock \showarticletitle{One-shot tuner for deep learning compilers}. In \bibinfo{booktitle}{\emph{Proceedings of the 31st ACM SIGPLAN International Conference on Compiler Construction}}. \bibinfo{pages}{89--103}.
\newblock


\bibitem[Ryu and Sung(2021)]%
        {ryu2021metatune}
\bibfield{author}{\bibinfo{person}{Jaehun Ryu} {and} \bibinfo{person}{Hyojin Sung}.} \bibinfo{year}{2021}\natexlab{}.
\newblock \showarticletitle{Metatune: Meta-learning based cost model for fast and efficient auto-tuning frameworks}.
\newblock \bibinfo{journal}{\emph{arXiv preprint arXiv:2102.04199}} (\bibinfo{year}{2021}).
\newblock


\bibitem[Sadeghi et~al\mbox{.}(2024)]%
        {10.1145/3665314.3670843}
\bibfield{author}{\bibinfo{person}{Mohammad~Erfan Sadeghi}, \bibinfo{person}{Arash Fayyazi}, \bibinfo{person}{Seyedarmin Azizi}, {and} \bibinfo{person}{Massoud Pedram}.} \bibinfo{year}{2024}\natexlab{}.
\newblock \showarticletitle{PEANO-ViT: Power-Efficient Approximations of Non-Linearities in Vision Transformers}. In \bibinfo{booktitle}{\emph{Proceedings of the 29th ACM/IEEE International Symposium on Low Power Electronics and Design}} (Newport Beach, CA, USA) \emph{(\bibinfo{series}{ISLPED '24})}. \bibinfo{publisher}{Association for Computing Machinery}, \bibinfo{address}{New York, NY, USA}, \bibinfo{pages}{1–6}.
\newblock
\showISBNx{9798400706882}
\urldef\tempurl%
\url{https://doi.org/10.1145/3665314.3670843}
\showDOI{\tempurl}


\bibitem[Schulman et~al\mbox{.}(2017)]%
        {schulman2017proximal}
\bibfield{author}{\bibinfo{person}{John Schulman}, \bibinfo{person}{Filip Wolski}, \bibinfo{person}{Prafulla Dhariwal}, \bibinfo{person}{Alec Radford}, {and} \bibinfo{person}{Oleg Klimov}.} \bibinfo{year}{2017}\natexlab{}.
\newblock \showarticletitle{Proximal policy optimization algorithms}.
\newblock \bibinfo{journal}{\emph{arXiv preprint arXiv:1707.06347}} (\bibinfo{year}{2017}).
\newblock


\bibitem[Tokdar and Kass(2010)]%
        {tokdar2010importance}
\bibfield{author}{\bibinfo{person}{Surya~T Tokdar} {and} \bibinfo{person}{Robert~E Kass}.} \bibinfo{year}{2010}\natexlab{}.
\newblock \showarticletitle{Importance sampling: a review}.
\newblock \bibinfo{journal}{\emph{Wiley Interdisciplinary Reviews: Computational Statistics}} \bibinfo{volume}{2}, \bibinfo{number}{1} (\bibinfo{year}{2010}), \bibinfo{pages}{54--60}.
\newblock


\bibitem[Vasilache et~al\mbox{.}(2018)]%
        {vasilache2018tensor}
\bibfield{author}{\bibinfo{person}{Nicolas Vasilache}, \bibinfo{person}{Oleksandr Zinenko}, \bibinfo{person}{Theodoros Theodoridis}, \bibinfo{person}{Priya Goyal}, \bibinfo{person}{Zachary DeVito}, \bibinfo{person}{William~S Moses}, \bibinfo{person}{Sven Verdoolaege}, \bibinfo{person}{Andrew Adams}, {and} \bibinfo{person}{Albert Cohen}.} \bibinfo{year}{2018}\natexlab{}.
\newblock \showarticletitle{Tensor comprehensions: Framework-agnostic high-performance machine learning abstractions}.
\newblock \bibinfo{journal}{\emph{arXiv preprint arXiv:1802.04730}} (\bibinfo{year}{2018}).
\newblock


\bibitem[Wang et~al\mbox{.}(2022)]%
        {wang2022automating}
\bibfield{author}{\bibinfo{person}{Huanting Wang}, \bibinfo{person}{Zhanyong Tang}, \bibinfo{person}{Cheng Zhang}, \bibinfo{person}{Jiaqi Zhao}, \bibinfo{person}{Chris Cummins}, \bibinfo{person}{Hugh Leather}, {and} \bibinfo{person}{Zheng Wang}.} \bibinfo{year}{2022}\natexlab{}.
\newblock \showarticletitle{Automating reinforcement learning architecture design for code optimization}. In \bibinfo{booktitle}{\emph{Proceedings of the 31st ACM SIGPLAN International Conference on Compiler Construction}}. \bibinfo{pages}{129--143}.
\newblock


\bibitem[Yu et~al\mbox{.}(2022)]%
        {yu2022surprising}
\bibfield{author}{\bibinfo{person}{Chao Yu}, \bibinfo{person}{Akash Velu}, \bibinfo{person}{Eugene Vinitsky}, \bibinfo{person}{Jiaxuan Gao}, \bibinfo{person}{Yu Wang}, \bibinfo{person}{Alexandre Bayen}, {and} \bibinfo{person}{Yi Wu}.} \bibinfo{year}{2022}\natexlab{}.
\newblock \showarticletitle{The surprising effectiveness of ppo in cooperative multi-agent games}.
\newblock \bibinfo{journal}{\emph{Advances in Neural Information Processing Systems}}  \bibinfo{volume}{35} (\bibinfo{year}{2022}), \bibinfo{pages}{24611--24624}.
\newblock


\bibitem[Yu and Zhu(2020)]%
        {yu2020hyper}
\bibfield{author}{\bibinfo{person}{Tong Yu} {and} \bibinfo{person}{Hong Zhu}.} \bibinfo{year}{2020}\natexlab{}.
\newblock \showarticletitle{Hyper-parameter optimization: A review of algorithms and applications}.
\newblock \bibinfo{journal}{\emph{arXiv preprint arXiv:2003.05689}} (\bibinfo{year}{2020}).
\newblock


\bibitem[Zhang et~al\mbox{.}(2022)]%
        {zhang2022harl}
\bibfield{author}{\bibinfo{person}{Zining Zhang}, \bibinfo{person}{Bingsheng He}, {and} \bibinfo{person}{Zhenjie Zhang}.} \bibinfo{year}{2022}\natexlab{}.
\newblock \showarticletitle{HARL: Hierarchical Adaptive Reinforcement Learning Based Auto Scheduler for Neural Networks}. In \bibinfo{booktitle}{\emph{Proceedings of the 51st International Conference on Parallel Processing}}. \bibinfo{pages}{1--13}.
\newblock


\bibitem[Zhou et~al\mbox{.}(2022)]%
        {zhou2022towards}
\bibfield{author}{\bibinfo{person}{Yanqi Zhou}, \bibinfo{person}{Xuanyi Dong}, \bibinfo{person}{Tianjian Meng}, \bibinfo{person}{Mingxing Tan}, \bibinfo{person}{Berkin Akin}, \bibinfo{person}{Daiyi Peng}, \bibinfo{person}{Amir Yazdanbakhsh}, \bibinfo{person}{Da Huang}, \bibinfo{person}{Ravi Narayanaswami}, {and} \bibinfo{person}{James Laudon}.} \bibinfo{year}{2022}\natexlab{}.
\newblock \showarticletitle{Towards the co-design of neural networks and accelerators}.
\newblock \bibinfo{journal}{\emph{Proceedings of Machine Learning and Systems}}  \bibinfo{volume}{4} (\bibinfo{year}{2022}), \bibinfo{pages}{141--152}.
\newblock


\end{thebibliography}
